\documentclass[10pt,twocolumn,letterpaper]{article}

\usepackage{cvpr}              
\definecolor{cvprblue}{rgb}{0.21,0.49,0.74}
\usepackage[pagebackref,breaklinks,colorlinks,allcolors=cvprblue]{hyperref}
\usepackage[accsupp]{axessibility}  
\usepackage{amsmath}
\usepackage{booktabs}
\usepackage{amssymb}
\usepackage{multirow}
\usepackage{standalone}
\usepackage{makecell}

\def\paperID{45974} 
\def\confName{CVPR}
\def\confYear{2026}

\title{GS-CLIP: Zero-shot 3D Anomaly Detection by Geometry-Aware Prompt and Synergistic View Representation Learning}

\author{
    Zehao Deng\textsuperscript{1,2}, 
    An Liu\textsuperscript{1}\thanks{Corresponding author.}\hspace{0.4em}, 
    Yan Wang\textsuperscript{2}\footnotemark[1] 
    \vspace{2mm} \\ 
    \textsuperscript{1}School of Computer Science and Technology, Soochow University \\
    \textsuperscript{2}Institute for AI Industry Research (AIR), Tsinghua University\\
    {\ttfamily\small 2327406010@stu.suda.edu.cn, anliu@suda.edu.cn, wangyan@air.tsinghua.edu.cn}
}

\begin{document}

\maketitle

\begin{abstract}
Zero-shot 3D Anomaly Detection is an emerging task that aims to detect anomalies in a target dataset without any target training data, which is particularly important in scenarios constrained by sample scarcity and data privacy concerns. While current methods adapt CLIP by projecting 3D point clouds into 2D representations, they face challenges. The projection inherently loses some geometric details, and the reliance on a single 2D modality provides an incomplete visual understanding, limiting their ability to detect diverse anomaly types. To address these limitations, we propose the Geometry-Aware Prompt and Synergistic View Representation Learning (GS-CLIP) framework, which enables the model to identify geometric anomalies through a two-stage learning process. In stage 1, we dynamically generate text prompts embedded with 3D geometric priors. These prompts contain global shape context and local defect information distilled by our Geometric Defect Distillation Module (GDDM). In stage 2, we introduce Synergistic View Representation Learning architecture that processes rendered and depth images in parallel. A Synergistic Refinement Module (SRM) subsequently fuses the features of both streams, capitalizing on their complementary strengths. Comprehensive experimental results on four large-scale public datasets show that GS-CLIP achieves superior performance in detection. Code can be available at \href{https://github.com/zhushengxinyue/GS-CLIP}{https://github.com/zhushengxinyue/GS-CLIP}.

\end{abstract}

\section{Introduction}

\begin{figure}[t]
    \centering
    \includegraphics[width=\columnwidth]{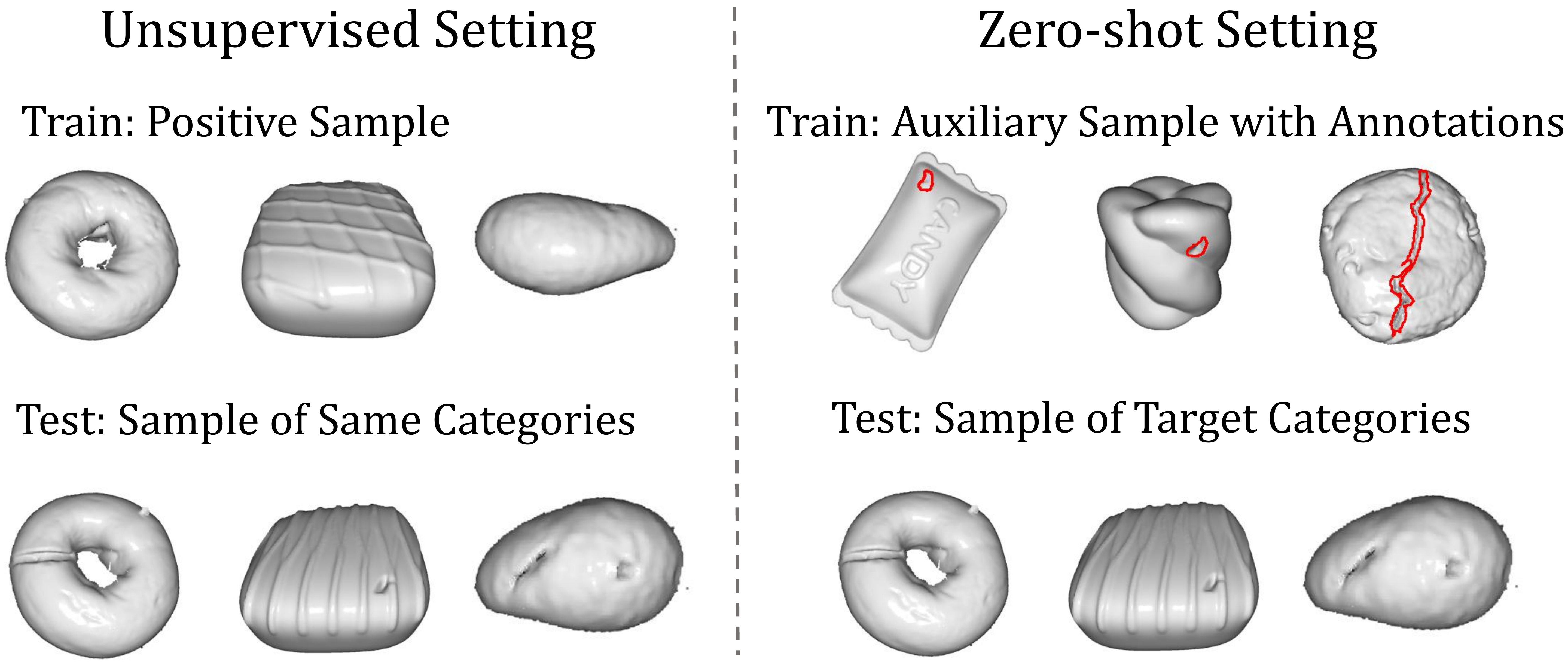}
    \caption{Comparison of task settings between traditional Unsupervised 3D Anomaly Detection (U3DAD) and Zero-shot 3D Anomaly Detection (ZS3DAD). U3DAD is trained on positive (normal) samples and tested on samples of the same categories; ZS3DAD is trained on auxiliary, annotated data and tested on unseen target categories.}
    \label{fig:paradigm_comparison}
\end{figure}

3D anomaly detection is a critical process in industrial manufacturing, playing a key role in ensuring the stable operation of industrial systems and reducing losses\cite{liu2024deep}. Traditional 3D anomaly detection relies on unsupervised methods \cite{bergmann2023anomaly,zhou2022pull,liu2023real3d,zhou2024r3d,wang2023multimodal,costanzino2024multimodal,gu2024rethinking,chen2023easynet,ye2024po3ad,zavrtanik2024cheating}, which require training on normal samples of the target category. However, acquiring sufficient training data for target categories is often difficult due to issues like commercial confidentiality and data privacy concerns \cite{gu2024filo, gu2024anomalygpt,jeong2023winclip}. To address these challenges, Zero-Shot 3D Anomaly Detection (ZS3DAD) has been proposed. As illustrated in Figure \ref{fig:paradigm_comparison}, the paradigm aims to train a general-purpose model using auxiliary data, without needing any samples from the target category.

Contrastive Language-Image Pre-training (CLIP) \cite{radford2021learning}, pre-trained on a massive dataset of hundreds of millions of image-text pairs, has established powerful joint visual-language representations. Consequently, CLIP has demonstrated strong generalization capabilities, particularly in Zero-shot 2D Anomaly Detection \cite{jeong2023winclip,gu2024anomalygpt,zhou2024anomalyclip,cao2024adaclip}. A few studies \cite{cheng2024mvp,zhou2024pointad} have already applied CLIP to ZS3DAD by projecting a 3D point cloud from multiple angles to obtain multi-view 2D images. After detecting and localizing anomalies in each 2D image using CLIP, the results are back-projected to the 3D point cloud. For example, PointAD \cite{zhou2024pointad} projects point clouds into rendered images and constructs a 3D representation from the resulting visual features to understand 3D anomalies from both point and pixel perspectives. Similarly, MVP-PCLIP \cite{cheng2024mvp} utilizes depth maps and learns visual and text prompts to help CLIP better understand anomalies. However, these methods still have two major limitations:

1) \textbf{Lack of 3D Geometric Structure Awareness.} The 3D-to-2D projection is lossy, compressing stereometric structures into planar pixels and inevitably discarding critical geometric details. Consequently, the model learns not the physical geometric form of anomalies, but their visual proxies in 2D images. When the visual features of a geometric anomaly are not prominent from a specific viewpoint, this indirect detection paradigm may be ineffective, significantly limiting the model's detection performance and generalization ability.

\begin{figure}[h!]
    \centering
    \includegraphics[width=\columnwidth]{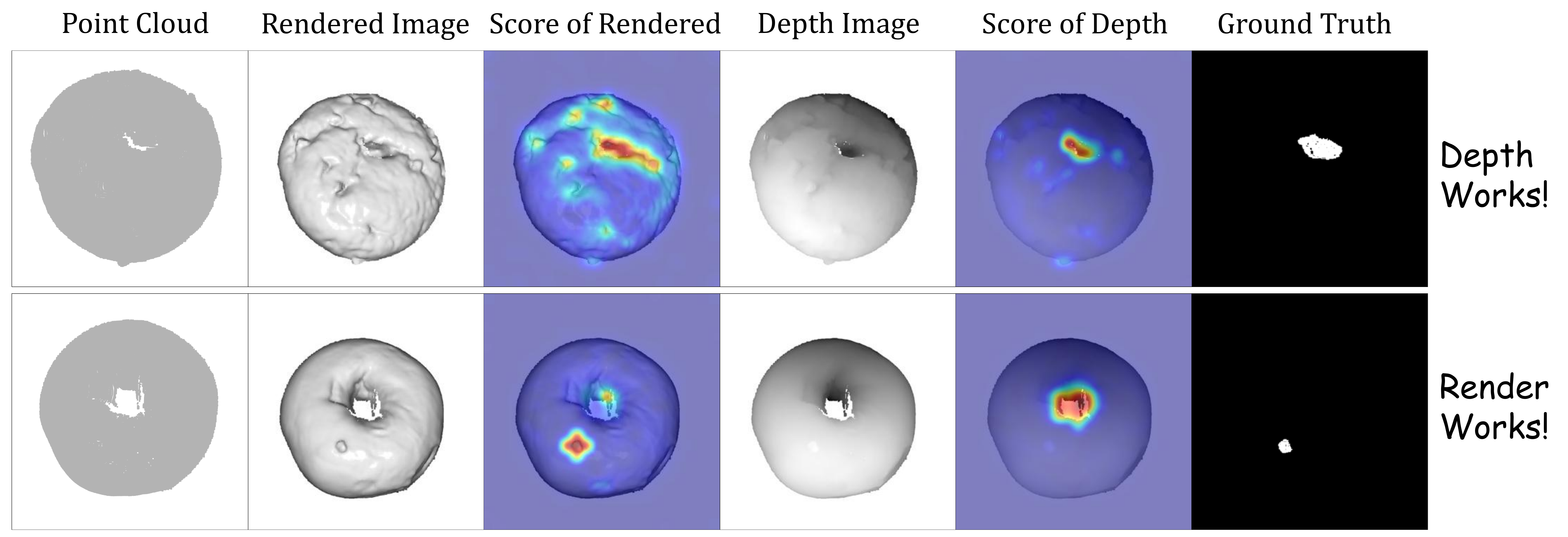}
    \caption{Example of the complementarity of rendered and depth images in anomaly detection. In the top row, the depth map effectively ignores surface texture interference to clearly show the dent anomaly; in the bottom row, the rendered image better captures the slight protrusion with insignificant depth change through lighting and shadow variations.}
    \label{fig:render_vs_depth}
\end{figure}

2) \textbf{Insufficient Utilization of Visual Information.} Current methods typically rely on a single type of 2D representation. However, as shown in Figure \ref{fig:render_vs_depth}, each representation has distinct pros and cons: rendered images are rich in appearance and texture information but are sensitive to lighting and rendering quality, and can sometimes introduce rendering artifacts that interfere with anomaly detection. In contrast, depth images better reflect the overall geometric structure, but fail to capture small details with minor depth variations. Therefore, reliance on a single image modality can limit both model performance and generalization capabilities.

To overcome these limitations, we propose Geometry-Aware Prompt and Synergistic View Representation Learning (GS-CLIP). In the first stage, we introduce \textbf{Geometry-Aware Prompt Learning} to optimize the text-end. This method trains a 3D feature extractor that not only injects the global features of the point cloud into the prompt as overall shape context but also identifies and synthesizes defect information from local geometric features via an \textbf{Geometric Defect Distillation Module (GDDM)}. This information is asymmetrically injected into the anomaly text prompt, providing the model with a direct geometric anomaly prior. This text generator dynamically creates prompts embedded with geometric structure information based on the 3D point cloud, helping to uncover subtle geometric anomalies in 2D images. In the second stage, we design a \textbf{Synergistic View Representation Learning} architecture. This architecture processes rendered and depth images in parallel. The rendering stream is fed directly into the vision encoder, while the depth image stream undergoes fine-tuning with LoRA to bridge the domain gap. The features extracted from both streams are then deeply fused through a \textbf{Synergistic Refinement Module (SRM)} to fully exploit the complementarity of the two visual representations. 

Our contributions can be summarized as follows:
\begin{itemize}
    \item We propose GS-CLIP, a framework that bridges the gap between 2D vision-language models and 3D anomaly detection. Our two-stage strategy enables CLIP to perceive and understand 3D structural anomalies from 2D multi-view images.
    \item We introduce Geometry-Aware Prompt Learning, which dynamically generates text prompts containing 3D geometric information to better reveal subtle geometric anomalies in 2D images.
    \item We design a Synergistic View Representation Learning architecture and introduce a Synergistic Refinement Module to effectively fuse complementary information from rendered and depth images.
    \item Experiments on four public datasets show that our method surpasses existing SOTA models on both object-level and point-level metrics.
\end{itemize}

\section{Related Work}

\subsection{Zero-shot 3D Anomaly Detection}
Zero-shot 3D anomaly detection aims to detect and localize anomalies without access to a target dataset, with its core challenge being the ability to learn general anomaly patterns that can generalize to unseen categories. The few existing works often leverage CLIP to address this problem \cite{cheng2024mvp,zhou2024pointad}. Their common pipeline involves projecting 3D data into multi-view 2D images, processing these images with CLIP, and subsequently back-projecting the 2D outputs onto the 3D point cloud. For instance, PointAD \cite{zhou2024pointad} utilizes a hybrid learning strategy that comprehends 3D anomalies from both point and pixel perspectives by back-projecting visual features from multi-view rendered images. In another work, MVP-PCLIP \cite{cheng2024mvp} uses depth maps and constructs key-layer visual prompts and adaptive text prompts to fine-tune CLIP. In contrast, 3DzAL \cite{wang2024towards} operates independently of CLIP, creating pseudo-anomalous training signals in 3D space to train a multi-branch network for identifying general geometric anomalies. Our work follows the CLIP-based multi-view projection paradigm, but distinguishes itself by integrating both depth and rendered images and by designing geometry-aware prompts to enhance CLIP's perception of 3D geometric structures.

\subsection{Unsupervised 3D Anomaly Detection}
Unsupervised Anomaly Detection is mainly divided into two categories. One category is based on feature embedding \cite{liu2023real3d,wang2023multimodal,rudolph2023asymmetric,costanzino2024multimodal,gu2024rethinking,zhu2024towards,liang2025look}, where the fundamental idea is to construct a distribution or boundary of normal samples and treat any deviations as anomalies, such as the teacher-student network in 3D-ST \cite{bergmann2023anomaly} and the memory bank in Reg3D-AD \cite{liu2023real3d}. The other category is based on reconstruction \cite{chen2023easynet,li2024towards,ye2024po3ad,zavrtanik2024cheating,zhou2024r3d}, where anomaly localization is performed using reconstruction errors. For instance, R3D-AD \cite{zhou2024r3d} designed an anomaly simulation strategy and used a diffusion model for reconstruction, and IMRNet \cite{li2024towards} employed iterative masked reconstruction with dual contrast in both point and feature spaces to derive anomaly scores. While these methods have achieved good performance, they depend on normal samples from the target category for training. When factors like data privacy and collection costs make it difficult to obtain even normal samples, the applicability of these methods is limited, which further motivates the exploration of less data-dependent and more generalizable learning paradigms, such as ZS3DAD.

\subsection{Zero-shot 2D Anomaly Detection}
Zero-shot 2D Anomaly Detection (ZS2DAD), much like ZS3DAD, is typically accomplished using CLIP. WinCLIP \cite{jeong2023winclip} was one of the first works to apply CLIP to ZS2DAD tasks. By designing various prompts and combining them with a multi-scale sliding window strategy, it demonstrates the immense potential of CLIP in ZSAD. Later, it was discovered that directly applying pre-trained CLIP or simple adaptations often leads to domain shift problems \cite{lin2025survey}. Consequently, many works \cite{chen2023aprilgan,cao2024adaclip,chen2024clipad,gu2024anomalygpt,li2024promptad,qu2024vcpclip,zhou2024anomalyclip,ma2025aaclip,qu2025bayesian, deng2025dde,sadikaj2025multiads} have designed different learnable modules for Parameter-Efficient Fine-Tuning (PEFT). Some methods use very small modules to adapt CLIP, such as AnomalyGPT \cite{gu2024anomalygpt} and APRIL-GAN \cite{chen2023aprilgan}, which both use linear layers for model alignment. Other methods focus on text prompts, for example, AnomalyCLIP \cite{zhou2024anomalyclip} uses general, learnable, object-agnostic text prompts, while AA-CLIP \cite{ma2025aaclip} creates text anchors to differentiate normal and abnormal semantics. Our work builds upon this prompt-centric approach, but uniquely leverages 3D information to construct geometry-aware prompts.

\begin{figure*}[t]
    \centering
    \includegraphics[width=0.9\textwidth]{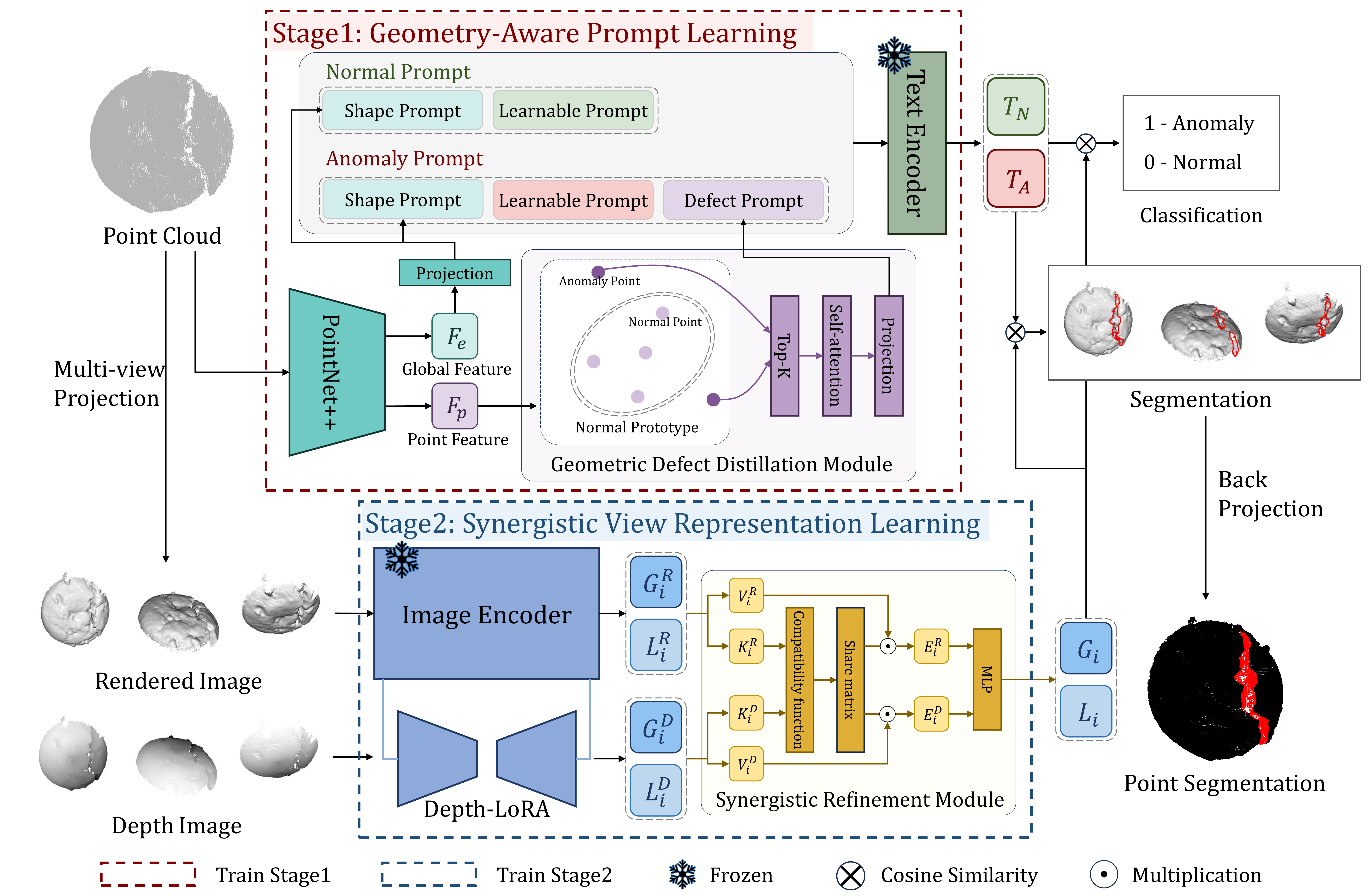}
    \caption{The overall architecture of GS-CLIP. The framework is optimized through a two-stage learning strategy. In stage 1, we generate text prompts embedded with geometric priors using a 3D feature extractor and a Geometric Defect Distillation Module. In stage 2, we design a synergistic architecture that processes rendered images and a LoRA-optimized depth image branch in parallel. The features from both branches are deeply fused by the Synergistic Refinement Module and finally compared with the text prompts to compute similarity for classification and segmentation.}
    \label{fig:framework}
\end{figure*}

\section{Methodology}
\subsection{Overview}
The architecture of our proposed model is illustrated in Figure \ref{fig:framework}. We employ a two-stage learning strategy to sequentially optimize the text and vision components of the framework.
\begin{itemize}
    \item \textbf{Stage 1: Geometry-Aware Prompt Learning.} In this stage, with the visual components frozen, we focus on training a text prompt generator. It dynamically creates shape prompts by extracting global geometric information from the 3D point cloud, and defect prompts by identifying local details via our proposed Geometric Defect Distillation Module. These text prompts, embedded with 3D structural priors, provide crucial cues for detecting and localizing anomalies from 2D projected images.
    \item \textbf{Stage 2: Synergistic View Representation Learning.} In this stage, the text prompt generator trained in Stage 1 is frozen, and our focus shifts to training the visual components. This stage features a dual-stream architecture: rendered images are processed by the original frozen vision encoder, while depth maps are passed through a parallel branch of the encoder fine-tuned with LoRA (Depth-LoRA). The features from both streams are then integrated by a Synergistic Refinement Module to facilitate information exchange and leverage their complementary strengths.
\end{itemize}
This two-stage strategy first ensures that the text prompt generator learns to robustly capture and describe 3D geometric anomalies. This, in turn, provides a high-quality, semantically rich optimization target for the vision-language alignment performed in the second stage.

\subsection{Geometry-Aware Prompt Learning}
\subsubsection{3D Feature Extraction and Shape Prompt}
For a given object's point cloud data $P \in \mathbb{R}^{n \times 3}$, where $n$ is the number of points, and each point has XYZ coordinates, we use a pre-trained PointNet++ \cite{qi2017pointnet++} ($\Psi_{pn}$) as our 3D feature extractor:
\begin{equation}
    F_p, F_e = \Psi_{pn}(P)
    \label{eq:pointnet}
\end{equation}
where $F_p \in \mathbb{R}^{n \times d_{pn}}$ represents the local geometric feature vector for each point, and $d_{pn}$ is the output dimension of PointNet++. $F_e \in \mathbb{R}^{d_e}$ is the global feature obtained by pooling the deep outputs of the PointNet++ encoder, representing an abstract understanding of the object's overall structure and shape, where $d_e$ is the encoder's output dimension.

To provide the text prompt with a macroscopic understanding of the object's overall shape, we project the global feature $F_e$ through a projection layer $\text{Proj}(\cdot)$ to obtain the overall shape prompt $t_s = \text{Proj}(F_e) \in \mathbb{R}^{d}$, where $d$ is the intrinsic dimension of CLIP.

\subsubsection{Geometric Defect Distillation Module}
We posit that an anomaly's essence lies in its deviation from the normal pattern \cite{yi2020patch}. Based on this, we design a normal prototype memory bank $\mathcal{P} = \{p_1, p_2, \dots, p_l\} \in \mathbb{R}^{l \times d_{pn}}$ consisting of $l$ learnable vectors. During training, these prototypes are implicitly guided to fit the distribution of normal local geometric features to minimize misclassification of normal samples, thereby forming a compact normal feature manifold.

Subsequently, for each point's local geometric feature $f_i \in \mathbb{R}^{d_{pn}}$ in $F_p$, we calculate a geometric outlier score $s_i$. This score is defined as the distance between the point feature and its most similar prototype in the memory bank:
\begin{equation}
    s_i = 1 - \max_{p_j \in \mathcal{P}} \frac{f_i \cdot p_j}{\|f_i\| \|p_j\|}
    \label{eq:outlier_score}
\end{equation}

Based on the calculated outlier scores, we select the features of the top-$k$ points with the highest scores, forming a feature set $\mathcal{F}_T = \{f_{T_1}, f_{T_2}, \dots, f_{T_k}\}$ that contains the most suspicious information. To capture the intrinsic structural relationships among these outlier points (e.g., whether they collectively form a scratch or a dent), we feed this subset into a self-attention-based aggregation network. This allows different outlier features to interact, distilling a holistic understanding of the entire defect region. Finally, through projection, we obtain the defect prompt $t_d \in \mathbb{R}^{k \times d}$:
\begin{equation}
    t_d = \text{Proj}(\text{SelfAttention}(\mathcal{F}_T)).
    \label{eq:defect_prompt}
\end{equation}

\subsubsection{Semantic Concatenation}
Finally, we concatenate the aforementioned geometric prompts to form the normal prompt $t_N$ and the anomaly prompt $t_A$:
\begin{align}
    t_N &= \text{Concat}(t_s, t_l) \label{eq:normal_prompt} \\
    t_A &= \text{Concat}(t_s, t_l, t_d) \label{eq:anomaly_prompt}
\end{align}
where $t_l \in \mathbb{R}^{q \times d}$ represents learnable prompts, and $q$ is the length of learnable prompts. These are passed through a frozen text encoder to obtain the normal text embedding $T_N \in \mathbb{R}^d$ and the anomaly text embedding $T_A \in \mathbb{R}^d$, which are used to compute similarity with visual features for anomaly detection.

\subsection{Synergistic View Representation Learning}
\subsubsection{Depth-LoRA}
RGB rendered images and their corresponding depth maps contain different information. Rendered images are rich in texture and detail, while depth maps better represent geometric information such as concavities and convexities. Therefore, we perform multi-view rendering and projection of the point cloud \cite{cao2024complementary} to obtain $v$ pairs of rendered images $\{I_i^R\}_{i=1}^v$ and depth maps $\{I_i^D\}_{i=1}^v$, with corresponding angles. To effectively process these two different inputs, we design a dual-stream architecture.

Since CLIP is trained on real images, it is naturally adapted to rendered images. Thus, we use a frozen, pre-trained Vision Transformer (ViT) to extract global features $G_i^R \in \mathbb{R}^d$ and local features $L_i^R \in \mathbb{R}^{p \times d}$ for the $i$-th view, where $d$ is CLIP's intrinsic dimension and $p$ is the number of patches.

For the depth map, we adopt the Low-Rank Adaptation \cite{hu2022lora}  (LoRA) technique. We selectively apply this adaptation rule to the linear layers of the MLP. The MLP in the original ViT can be represented as:
\begin{equation}
    \text{MLP}(x)=W_2 \cdot \text{GELU}(W_1x),
\end{equation}
where $W_1$ and $W_2$ represent the linear transformation matrices. The MLP after applying LoRA can be represented as:
\begin{equation}
    x'=\text{GELU}(W_1x+\gamma B_1A_1x), 
\end{equation}
\begin{equation}
    \text{MLP}'(x)=W_2x'+ \gamma B_2A_2x'.
\end{equation}
    
Through this method, we fine-tune only the MLP layers to adapt to the feature distribution of the depth map, while completely preserving the powerful spatial relationship modeling capability of the pre-trained model in the Multi-Head Self-Attention block.

\subsubsection{Synergistic Refinement Module}
To effectively combine features from rendered and depth images, we introduce a Synergistic Refinement Module. It receives global features $G_i^R \in \mathbb{R}^d$ and local features $L_i^R \in \mathbb{R}^{p \times d}$ from the rendered image, and global features $G_i^D \in \mathbb{R}^d$ and local features $L_i^D \in \mathbb{R}^{p \times d}$ from the depth map. The fusion of local features follows the same process as global features. Taking global features as an example, it first generates two key-value pairs: $K_i^R, V_i^R$ and $K_i^D, V_i^D$. Then, a shared matrix is generated through a compatibility function $f$:
\begin{equation}
    S = f(K_i^R, K_i^D) = f_1(K_i^R) \times f_2(K_i^D)^T,
    \label{eq:SRM_s}
\end{equation}
where we choose bidirectional multiplicative attention as the compatibility function, and $f_1$ and $f_2$ are linear projections. Based on this shared matrix, we compute two sets of attention weights and perform information aggregation:
\begin{equation}
    E_i^R, E_i^D = \text{Softmax}(S) \cdot V_i^R, \text{Softmax}(S^T) \cdot V_i^D.
    \label{eq:SRM_e}
\end{equation}
Finally, these two enhanced features are concatenated and fused through a small MLP network to obtain a synergistic global feature representation for both modalities:
\begin{equation}
    G_i = \text{MLP}(\text{Concat}(E_i^R, E_i^D)).
    \label{eq:SRM_l}
\end{equation}

\subsection{Anomaly Score Map}
Both visual and text features are mapped into a unified vector space by pre-trained CLIP encoders. Therefore, the overall classification of an image is determined by computing the similarity between the global visual feature $G_i$ and the text features $T_A, T_N$:
\begin{equation}
    \hat{y_i} = \frac{\exp(\langle G_i, T_A \rangle / \tau)}{\exp(\langle G_i, T_N \rangle / \tau) + \exp(\langle G_i, T_A \rangle / \tau)},
    \label{eq:classification_prob}
\end{equation}
where $\hat{y_i}$ is the probability that the $i$-th view image is anomalous, $\langle \cdot, \cdot \rangle$ denotes cosine similarity, and $\tau$ is a temperature coefficient. The final anomaly probability $\hat{y}$ for the 3D point cloud is aggregated from all views: $\hat{y} = \frac{1}{v} \sum_{i=1}^v \hat{y_i}$.

Simultaneously, an anomaly score map is obtained by aligning the local visual features $L_i$:
\begin{align}
    M_i^N &= \text{Up}\left(\frac{\exp(\langle L_i, T_N \rangle)}{\exp(\langle L_i, T_N \rangle) + \exp(\langle L_i, T_A \rangle)}\right), \label{eq:map_normal} \\
    M_i^A &= \text{Up}\left(\frac{\exp(\langle L_i, T_A \rangle)}{\exp(\langle L_i, T_N \rangle) + \exp(\langle L_i, T_A \rangle)}\right), \label{eq:map_anomaly}
\end{align}
where $M_i^N \in \mathbb{R}^{h \times w}$ is the normal score map and $M_i^A \in \mathbb{R}^{h \times w}$ is the anomaly score map, $\text{Up}(\cdot)$ is bilinear interpolation upsampling, and $h, w$ are the height and width of the image. The anomaly score map for the $i$-th view $ M_i = G_{\sigma}(\frac{1}{2}(\mathbf{I} - M_i^N) + \frac{1}{2}M_i^A)$, where $G_{\sigma}$ is a Gaussian filter and $\mathbf{I}$ is a matrix of all ones.

To back-project the scores to the 3D point cloud and obtain a score for each point, we follow the approach of PointAD. During the projection from 3D point cloud to 2D images, we record the occlusion status of each point. Specifically, $H_i \in \mathbb{R}^n$ indicates whether each point in the 3D cloud is visible in the $i$-th view (1 for visible, 0 for occluded). Thus, the 3D anomaly score maps are:
\begin{equation}
    M = \frac{1}{v} \sum_{i=1}^v \left(R_i^{-1}(M_i) \circ H_i\right), 
\end{equation}
where $M \in \mathbb{R}^{n}, $ $R_i^{-1}$ is the back-projection matrix for the $i$-th view, and $\circ$ represents element-wise multiplication.

\begin{table*}[t]
\centering
\small
\setlength{\tabcolsep}{1.7mm}
\caption{Zero-shot 3D anomaly detection results. O-AUROC (O-R), O-AP (O-A) at object-level and P-AUROC (P-R), P-PRO (P-P) at point-level metrics are presented. Best results are in \textbf{bold}, second-best are \underline{underlined}.}
\label{tab:zs3dad_results}
\resizebox{\textwidth}{!}{
\begin{tabular}{@{}lc cc cc cc cc@{}}
\toprule
\multirow{2}{*}{Model}& \multirow{2}{*}{Public} &\multicolumn{2}{c}{MVTec3D-AD} & \multicolumn{2}{c}{Eyecandies} & \multicolumn{2}{c}{Real3D-AD} & \multicolumn{2}{c}{Anomaly-ShapeNet} \\
\cmidrule(lr){3-4} \cmidrule(lr){5-6} \cmidrule(lr){7-8} \cmidrule(lr){9-10}
& & (O-R, O-A) & (P-R, P-P) & (O-R, O-A) & (P-R, P-P) & (O-R, O-A) & (P-R, P-P) & (O-R, O-A) & (P-R, P-P) \\
\midrule
\multicolumn{7}{l}{\textit{one-vs-rest setting}} \\
CLIP* & ICML'21 & (61.2, 85.8) & (80.3, 54.4) & (66.7, 69.2) & (81.2, 37.9) & (68.8, 72.3) & (45.9, -) & (78.2, 79.5) & (64.9, -) \\
AA-CLIP* & CVPR'25 & (74.2, 88.5) & (87.1, 61.6) & (66.5, 68.0) & (85.9, 39.3) & (74.8, 76.3) & (51.2, -) & (83.1 84.5) & (60.7, -) \\
3DzAL & WACV'25 & (53.1, -) & (70.8, -) & (55.8, -) & (72.7, -) & (52.4, -) & (67.5, -) & (58.6, -) & (66.1, -) \\
MVP-PCLIP & arXiv'24 &(81.3, 92.7) & (94.6, 83.6) & (69.3, 72.7) & (90.8, 67.8) & (\underline{74.9}, 75.3) & (\underline{75.9}, --) & (78.7, 83.8) & (70.4, -) \\
PointAD & NeurIPS'24 & (\underline{82.0}, \underline{94.2}) & (\underline{95.5}, \underline{84.4}) & (\underline{69.1}, \underline{73.8}) & (\underline{92.1}, \underline{71.3}) & (74.8, \underline{76.9}) & (73.5, --) & (\underline{82.6}, \underline{85.6}) & (\underline{74.1}, -) \\
GS-CLIP & - & (\textbf{83.6}, \textbf{96.5}) & (\textbf{96.3}, \textbf{86.4}) & (\textbf{71.5}, \textbf{75.9}) & (\textbf{93.1}, \textbf{73.8}) & (\textbf{76.4}, \textbf{77.7}) & (\textbf{76.3}, --) & (\textbf{84.1}, \textbf{86.8}) & (\textbf{75.2}, -) \\
\midrule
\multicolumn{7}{l}{\textit{cross-dataset setting}} \\
AA-CLIP* & CVPR'25 & - & - & (63.6, 65.3) & (82.6, 39.8) & (70.2, 71.7) & (55.2, -) & (76.2, 77.5) & (61.8, -) \\
MVP-PCLIP & arXiv'24 & - & - &(66.7, 70.7) & (88.3, 66.0) & (74.9, 75.6) & (\underline{73.9}, -) & (78.4, 82.5) & (71.1, -) \\ 
PointAD & NeurIPS'24 & - & - & (\underline{69.5}, \underline{74.3}) & (\underline{91.8}, \underline{71.4}) & (\underline{75.9}, \textbf{78.9}) & (71.6, -) & (\underline{82.3}, \textbf{85.6}) & (\underline{74.3}, -) \\
GS-CLIP & - & - & - & (\textbf{70.3}, \textbf{75.3}) & (\textbf{92.9}, \textbf{73.4}) & (\textbf{76.2}, \underline{77.1}) & (\textbf{74.7}, -) & (\textbf{82.6}, \underline{84.8}) & (\textbf{75.4}, -) \\
\bottomrule
\end{tabular}
}
\end{table*}

\subsection{Loss Function}
We use a binary cross-entropy classification loss $L_{cla}$, a segmentation loss $L_{seg}$ composed of Dice loss and Focal loss for both 2D and 3D outputs, as:
\begin{align}
    L_{cla} &= \text{BCE}(y, \hat{y}), \label{eq:loss_cla} \\
    L_{seg} &= \text{Dice}(M, Y) + \text{Focal}(M, Y) + \notag \\
            &\quad + \frac{1}{v} \sum_{i=1}^v (\text{Dice}(M_i, Y_i) + \text{Focal}(M_i, Y_i)). \label{eq:loss_seg}
\end{align}
where $y$ is object label, $Y$ represents the point-wise labels and $Y_i$ represents the labels for the $i$-th view.

Furthermore, we adds the \textbf{cross-view consistency loss} $L_{con}$. This loss encourages the model to learn a view independent global representation of the essence of the object, enhancing its generalization ability to any unknown perspective. First, we compute the average global feature for each object, $\bar{G} = \frac{1}{v}\sum_{i=1}^v G_i$, and then calculate the deviation of each view from this average:
\begin{equation}
    L_{con} = 1 - \frac{1}{v} \sum_{i=1}^v (\langle G_i, \bar{G} \rangle).
    \label{eq:loss_con}
\end{equation}

The total losses for the first and second stages are respectively with weights $\alpha$:
\begin{align}
    L_{stage1} &= L_{cla} +  L_{seg}, \label{eq:loss_s1} \\
    L_{stage2} &= L_{cla} +  L_{seg} + \alpha L_{con}. \label{eq:loss_s2}
\end{align}

\section{Experiments}
\subsection{Experimental Setup}
\paragraph{Datasets.} We conduct experiments on four datasets: MVTec3D-AD \cite{bergmann2022mvtec}, Real3D-AD \cite{liu2023real3d}, Eyecandies \cite{bonfiglioli2022eyecandies}, and Anomaly-ShapeNet \cite{li2024towards}. MVTec3D-AD and Real3D-AD are real-world datasets collected with industrial-grade structured-light 3D scanners, containing 10 and 12 categories of high-resolution objects, respectively, with all anomaly samples having precise point-wise annotations. Eyecandies and Anomaly-ShapeNet are synthetic datasets with 10 and 40 object categories, respectively. Additionally, MVTec3D-AD and Eyecandies include corresponding RGB images for multimodal anomaly detection tasks. we use the common zero-shot setting: one-vs-rest and cross-dataset. For one-vs-rest, we independently train on three selected categories for each of the four datasets and report the average evaluation results. For cross-dataset, we used a single category in MVTec3D-AD dataset as the sole auxiliary training data and tested on the other datasets. 

\paragraph{Evaluation Metrics.} The metrics we evaluate include Area Under the Receiver Operating Characteristic curve (AUROC), Average Precision (AP), and Per-Region Overlap (PRO). At the object-level, we use AUROC and AP, denoted as O-AUROC (O-R) and O-AP (O-A), respectively. At the point-level, we use AUROC and PRO, denoted as P-AUROC (P-R) and P-PRO (P-P), respectively.

\paragraph{Implementation Details and Baseline.} For the public CLIP model, we use the ViT-L/14@336px version as pre-trained weights, with all parameters frozen. Point clouds and 2D images are resized to 336x336. For most categories, we generate 9-view 2D images using the Open3D library by rotating the point cloud around the X-axis at angles of $\{\frac{4}{5}\pi, \frac{3}{5}\pi,..., -\frac{4}{5}\pi\}$. In the first training stage, we do not include the depth map, Depth-LoRA, or the SRM, using only the visual features from the render image. $\alpha$ is set to 1, the number of anomaly point features $k$ is set to 12, and the LoRA rank $r$ is set to 8. We use the Adam optimizer, training for 15 epochs in the first stage with a learning rate of 0.002, and for 10 epochs in the second stage with a learning rate of 0.0005, on a single NVIDIA RTX 3090 24GB GPU. For the original CLIP model and the SOTA model AA-CLIP in ZS2DAD, we simply used multi-view projection image to adapt it to ZS3DAD. Appendix B provide more details.

\subsection{Main Results}
\paragraph{Quantitative Comparison.}Table \ref{tab:zs3dad_results} shows the zero-shot performance on four large-scale datasets under the one-vs-rest setting. Compared to recent state-of-the-art methods, GS-CLIP achieves superior performance at both object and point levels. 
Compared to the second-best PointAD, GS-CLIP shows consistent improvements across the board, with an average increase of 1.8\% in O-AUROC, 1.6\% in O-AP, and 2.5\% in P-PRO. Vanilla CLIP and the adapted 2D SOTA method AA-CLIP yield modest performance, reflecting the inadequacy of applying 2D models directly to the 3D domain. The non-CLIP-based method 3DzAL also achieves limited results, possibly due to its failure to utilize effective anomaly signals. PointAD's reliance solely on rendered images makes it susceptible to lighting variations and rendering artifacts, which can obscure purely geometric anomalies. Conversely, MVP-PCLIP's dependence on depth maps, while strong for capturing overall shape, limits its ability to detect subtle surface defects like scratches or slight protrusions that cause minimal depth changes.
In contrast, the proposed GS-CLIP fundamentally enhances the model's perception by synergistically fusing both rendered and depth views while injecting explicit 3D geometric priors, thus outperforming other methods.

Furthermore, GS-CLIP demonstrates exceptional generalization. Cross-dataset setting is a test that can better evaluate the generalization ability of the model, as datasets involve significant shifts in object semantics, background, and data acquisition methods (real-world or synthetic). As shown in Table \ref{tab:zs3dad_results},  GS-CLIP achieves superior performance in cross-dataset setting. At the same time, compared with the performance under the one-vs-rest setting of the same dataset, the performance of the model has only a small decline.

\begin{figure}[h!]
    \centering
    \includegraphics[width=\columnwidth]{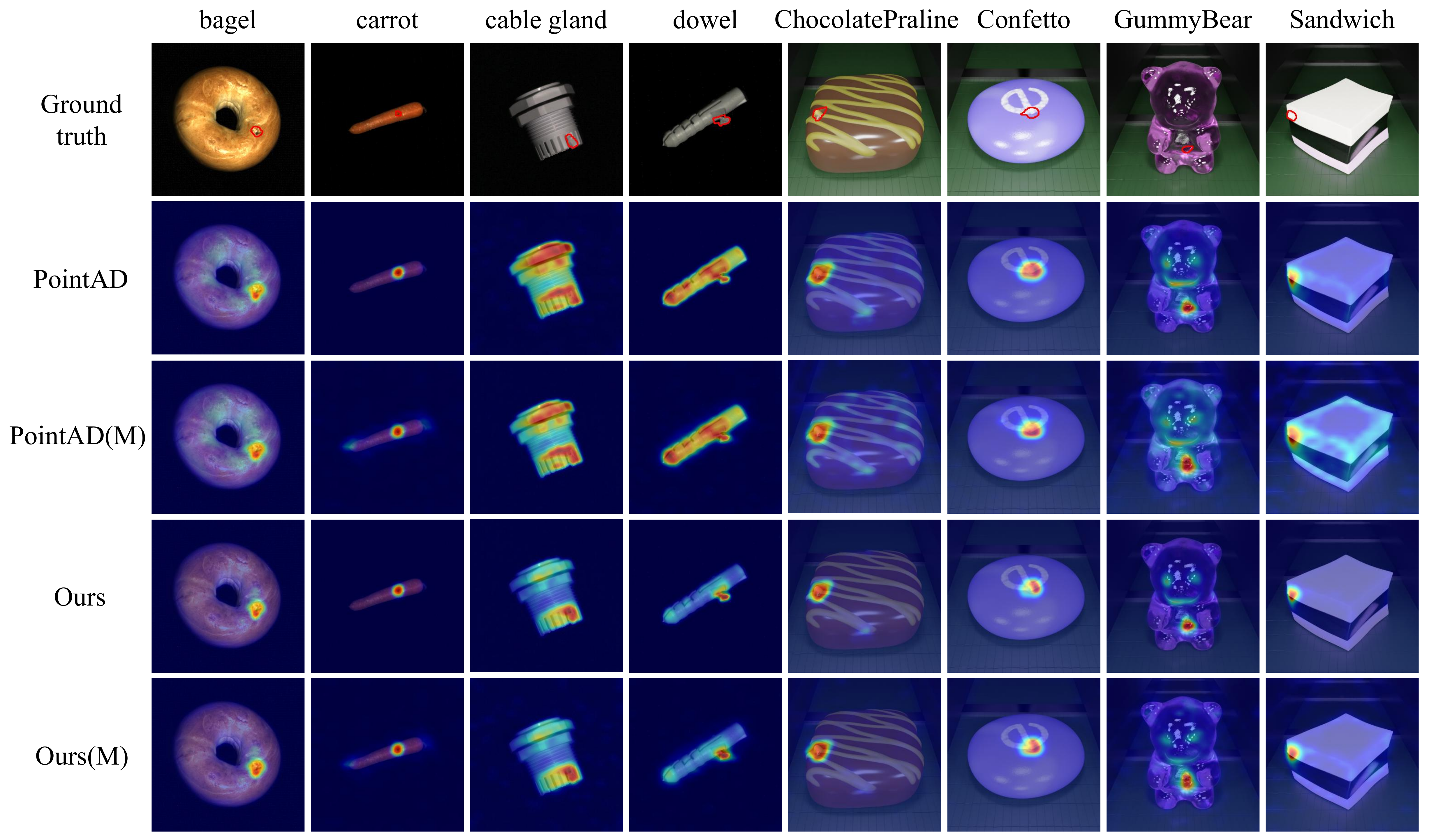}
    \caption{Qualitative comparison of anomaly score map between PointAD and our method. (M) represents multimodal, which is the result of integrating RGB images.}
    \label{fig:visualization}
\end{figure}

\paragraph{Qualitative Comparison.} As shown in Figure \ref{fig:visualization}, we visualize the anomaly score maps for MvTec3D-AD and Eyecandies. Our method achieves more precise segmentation and effectively suppresses anomaly scores in normal regions, preventing mis-detections. Especially for some objects with uneven surfaces, such as cable glands and dowels, our method can still identify anomalies accurately. This demonstrates the effectiveness and superiority of our method in dynamically generating shape and defect prompts.

\paragraph{Multimodal Results.} To further validate our model, for datasets that include corresponding RGB images with color, we adopt a plug-and-play multimodal fusion method \cite{zhou2024pointad}. As shown in Table \ref{tab:zsm3dad_results}, in the multimodal setting, the performance of all methods improves. RGB images provide real color and fine texture information that is missing in rendered and depth maps, which is crucial for detecting surface scratches, stains, and discolorations. Our method also achieves the best performance, with O-AUROC reaching 88.2\% on MVTec3D-AD and 79.3\% on Eyecandies, demonstrating the generalization and robustness of our method.

\begin{table}[h!]
\centering
\small
\caption{Multimodal zero-shot 3D anomaly detection results. Best results are in \textbf{bold}, second-best are \underline{underlined}.}
\label{tab:zsm3dad_results}
\resizebox{\columnwidth}{!}{
\begin{tabular}{@{}c cc cc@{}}
\toprule
\multirow{2}{*}{\textbf{Model}} & \multicolumn{2}{c}{\textbf{MVTec3D-AD}} & \multicolumn{2}{c}{\textbf{Eyecandies}} \\
\cmidrule(lr){2-3} \cmidrule(lr){4-5}
& (O-R, O-A) & (P-R, P-P) & (O-R, O-A) & (P-R, P-P) \\
\midrule
CLIP* & (60.4, 86.4) & (81.7, 56.0) & (73.0, 73.9) & (78.0, 31.8) \\
AA-CLIP* & (67.2, 89.4) & (94.5, 74.6) & (71.7, 69.1) & (86.8 74.2) \\
MVP-PCLIP & (85.7, 95.8) & (96.5, \underline{90.4}) & (\underline{78.2}, 80.1) & (94.9, 83.6) \\
PointAD   & (\underline{86.9}, \underline{96.1}) & (\underline{97.2}, 90.2) & (77.7, \underline{80.4}) & (\underline{95.3}, \underline{84.3}) \\
GS-CLIP   & (\textbf{88.2}, \textbf{97.5}) & (\textbf{97.6}, \textbf{91.3}) & (\textbf{79.3}, \textbf{82.2}) & (\textbf{95.8}, \textbf{86.4}) \\
\bottomrule
\end{tabular}
}
\end{table}

\subsection{Complexity Analysis}

We compare the computation overhead, including inference time per image, frames per second (FPS), and GPU memory usage. As shown in Table \ref{tab:complexity}, although our model’s inference time and memory usage are slightly higher than the others, its accuracy achieves SOTA across all four datasets. This significant accuracy gain demonstrates the effectiveness and superiority of our model design, achieving a better accuracy-efficiency trade-off in a multi-dimensional performance evaluation.

\begin{table}[h!]
\centering
\small 
\caption{Comparison of computation overhead with SOTA approaches on MVTec3D-AD.}
\label{tab:complexity}
\begin{tabular}{@{}cccc@{}}
\toprule
\textbf{Model} & \textbf{Time(s)} & \textbf{FPS} & \textbf{Memory Usage(MB)} \\ \midrule
CLIP* & 0.23 & 4.34 & 3312 \\
AA-CLIP* & 0.29 & 3.45 & 3826  \\
MVP-PCLIP  & 0.46 & 2.17 & 4583  \\
PointAD & 0.40 & 2.52 & 4275  \\
Ours & 0.51 & 1.96 & 5872 \\ \bottomrule
\end{tabular}
\end{table}

\subsection{Ablation Study}
\paragraph{Ablation of Key Modules.} To investigate the impact of key modules on the overall model, we conducted an ablation study by progressively adding components on MVTec3D-AD, with results shown in Table \ref{tab:ablation}. We first evaluated the performance using only a single 2D representation with learnable prompt. When we fused the two visual representations with SRM, all metrics improved significantly. On top of the two representative fusion, we further added the Shape Prompt and Defect Prompt respectively to the text prompt. As can be seen, object-level metrics improved further when adding Shape Prompt, demonstrating that providing macroscopic geometric context to the text prompts helps the model's overall judgment. Simultaneously, model's point-level metrics saw the largest increase when adding Defect Prompt and GDDM, with P-PRO rising from 84.8\% to 85.6\%. This proves that extracting and injecting specific defect descriptions from outlier points greatly enhances the model's ability to precisely localize anomalies. When the two prompts are integrated, the model effect is significantly improved compared with the single prompt. Finally, with the cross-view consistency loss $L_{con}$ of our design, the effect of the model is further improved to achieve the best.

\begin{table}[h!]
\centering
\small
\caption{Ablation study of key modules. SRM: Synergistic Refinement Module, SP: Shape Prompt, DP: Defect Prompt.}
\resizebox{\columnwidth}{!}{
\begin{tabular}{@{}cccccc@{}}
\toprule
 \textbf{SRM}& \textbf{SP} & \textbf{DP} & \textbf{$L_{con}$} & \textbf{ (O-R, O-A)} & \textbf{(P-R, P-P)} \\
\midrule
$\times$(renderings)& $\times$ &$\times$  &$\times$ & (80.9, 91.7) & (93.5, 83.1) \\
$\times$(depth map)& $\times$ &$\times$  &$\times$ & (81.4, 91.4) & (92.2, 82.5) \\
 $\checkmark$ & $\times$ & $\times$ &$\times$ & (82.3, 93.9) & (94.6, 84.8) \\
 $\checkmark$ & $\checkmark$ & $\times$ &$\times$ & (82.5, 94.8) & (95.2, 85.1) \\
 $\checkmark$ & $\times$ & $\checkmark$ &$\times$ & (82.9, 94.4) & (95.6, 85.6) \\
  $\checkmark$ & $\checkmark$ & $\checkmark$ & $\times$ & (\underline{83.1}, \underline{96.2}) & (\underline{96.0}, \underline{86.2}) \\
 $\checkmark$ & $\checkmark$ & $\checkmark$ & $\checkmark$ & \textbf{(83.6, 96.5)} & \textbf{(96.3, 86.4)} \\
\bottomrule
\end{tabular}
}
\label{tab:ablation}
\end{table}

\begin{figure}[h!]
    \centering
    \includegraphics[width=\columnwidth]{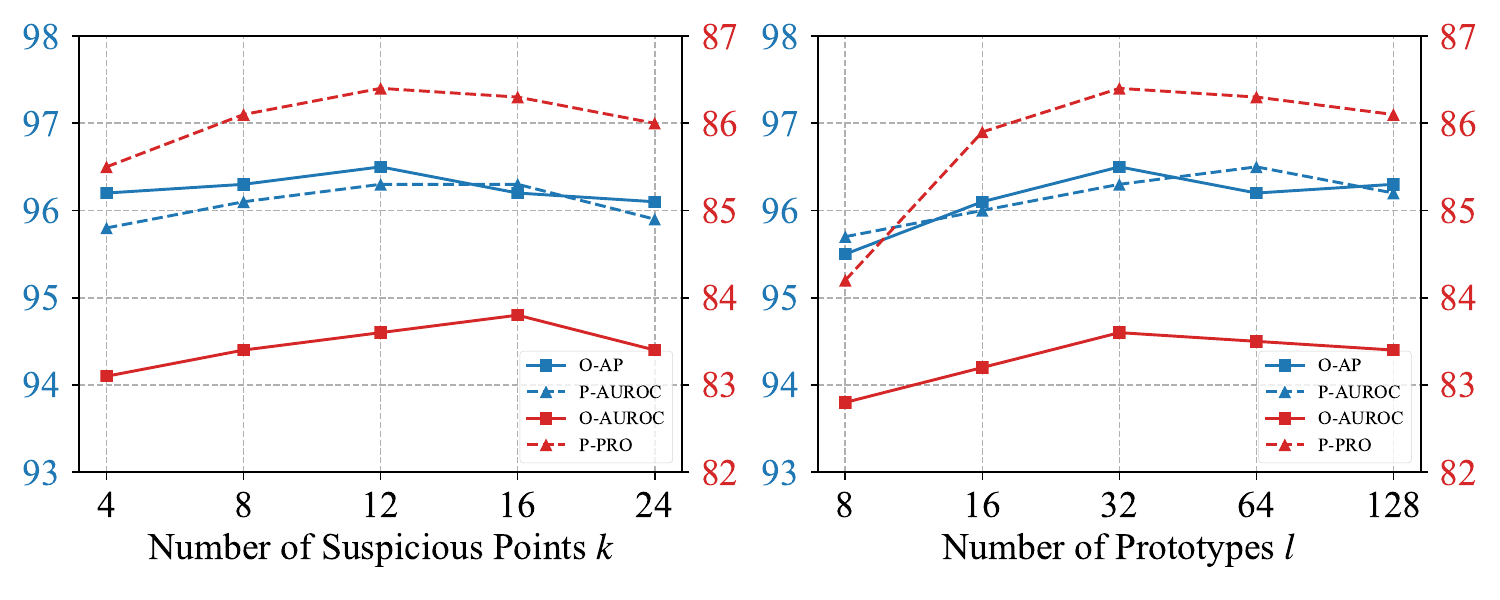}
    \caption{Parameters in GDDM.}
    \label{fig:GDDM1}
\end{figure}

\paragraph{Parameters in GDDM.}We investigated the influence of the number of candidate outlier points $k$ and the number of prototypes $l$ in GDDM, and the results are shown in Figure \ref{fig:GDDM1}. We observed that as $k$ increased from 4 to 12, the performance indicators of the model, especially the P-PRO, steadily improved from 85.5 to a peak of 86.4. This indicates that a larger $k$ value can enable the GDDM to capture a more complete representation of anomalous regions. However, as $k$ continued to increase. We believe this is because excessive $k$ values introduce normal point features as noise, thereby contaminating the purity of defect-specific features. 

As for number of prototypes $l$, the model performance is suboptimal when $l$ is small, which is likely due to the insufficient number of prototypes being unable to fully model normal sample features. As $l$ increases to 32, performance reaches its peak. If the $l$ continues to increase, the performance does not bring significant benefits or even decline. So we consider $l$=32 is sufficient to construct a representative and not overly complex normal feature manifold for the model.

\begin{figure}[h!]
    \centering
    \includegraphics[width=\columnwidth]{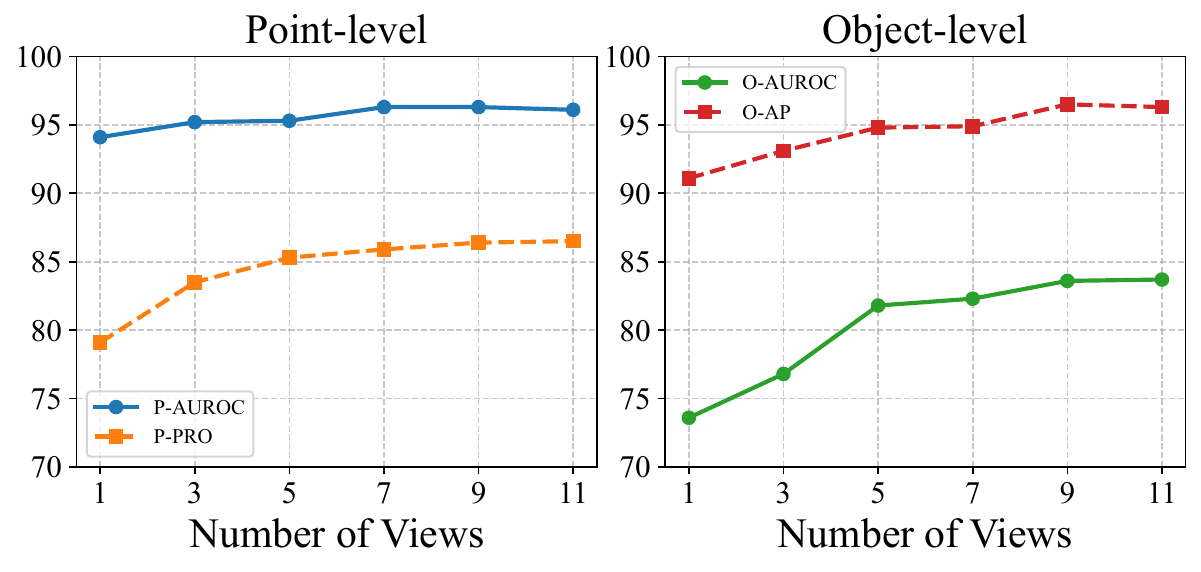}
    \caption{Ablation of Number of Views.}
    \label{fig:view_number}
\end{figure}

\paragraph{Number of Views.}Different views are crucial for understanding 3D anomaly, and having multiple views means a more comprehensive understanding of 3D anomaly. As shown in Figure \ref{fig:view_number}, with the increase of the number of views, the indicator shows an overall upward trend, reaching saturation at approximately 9 views. Regarding rendering conditions, prototype library size, hyperparameters, and other ablation experiments, please refer to the Appendix.

\section{Conclusion}
This paper proposes Geometry-Aware Prompt and Synergistic View Representation Learning (GS-CLIP) for ZS3DAD. The core of our framework is a LoRA-enhanced parallel visual encoder that synergistically processes render and depth images, and a geometry-aware prompt generator that injects global shape and local defect information into text prompts, achieving true geometry-aware semantic understanding. Extensive experiments on four public datasets demonstrate that our method significantly surpasses existing SOTA models on both object-level and point-level metrics. Exploring more direct 3D native representations and modal fusion methods may be worthy research directions for the future.

\section*{Acknowledgment}
This work is supported by Project Funded by Priority Academic Program Development of Jiangsu Higher Education Institutions; Wuxi Research Institute of Applied Technologies, Tsinghua University under Grant 20242001120; Hui-Chun Chin and Tsung-Dao Lee Chinese Undergraduate Research Endowment (CURE).

{
    \small
    \bibliographystyle{ieeenat_fullname}
    \bibliography{main}
}


\end{document}